\documentclass{article}
\usepackage[preprint]{spconf}
\usepackage{fancyhdr}

\usepackage{spconf}
\usepackage{graphicx}
\usepackage{color}
\usepackage{soul}
\definecolor{light-gray}{gray}{0.85}

\definecolor{silver}{rgb}{0.94,1,1}

\usepackage[linesnumbered,ruled,noend,noline]{algorithm2e} 
\usepackage{cite} 
\usepackage{url}  
\usepackage{ifthen}  
\usepackage{multicol}   
\usepackage{graphicx,courier,amssymb,amsmath,times}
\usepackage{epsfig,psfrag}
\usepackage{bm,graphicx,hhline,amsmath,amssymb,array}
\usepackage{rotating}
\usepackage{amsfonts}
\usepackage{amsmath}

\usepackage[implicit=false, bookmarks=false]{hyperref} 

\usepackage[all]{nowidow}
\usepackage{cases}
\usepackage{textcomp}
\usepackage{upgreek}
\usepackage{boldline}
\usepackage{booktabs}
\usepackage{comment}
\usepackage[nolist]{acronym}
\usepackage[caption=false, textfont=large]{subfig} 
\usepackage{orcidlink}

\usepackage{multirow}
\usepackage{tabularx} 
\newcolumntype{Y}{>{\centering\arraybackslash}X}

\usepackage{floatrow}
\definecolor{light-gray}{gray}{0.85}

\newcommand{\G}{\mathcal{G}}
\newcommand{\U}{\mathcal{U}}
\newcommand{\B}{B}

\title{MADRL-BASED UAVS TRAJECTORY DESIGN WITH ANTI-COLLISION MECHANISM IN VEHICULAR NETWORKS}
\name{{Leonardo Spampinato$^{1,2}$, Enrico Testi$^{1,2}$, Chiara Buratti$^{1,2}$, Riccardo Marini$^{1}$\orcidlink{0000-0003-1559-2603}}
\address{$^{1}$Wilab, CNIT, Bologna, Italy; \\ 
$^{2}$DEI Department, University of Bologna, Bologna, Italy; \\
\{leonardo.spampinato, enrico.testi, c.buratti\}@unibo.it, riccardo.marini@cnit.it}}

\begin{document}

\ninept
\maketitle

\thispagestyle{fancy}
\fancyhead{}
\lfoot{\copyright2024 IEEE. Personal use of this material is permitted. Permission from IEEE must be obtained for all other uses, in any current or future media, including reprinting/republishing this material for advertising or promotional purposes, creating new collective works, for resale or redistribution to servers or lists, or reuse of any copyrighted component of this work in other works.}

\begin{abstract}
In upcoming 6G networks, unmanned aerial vehicles (UAVs) are expected to play a fundamental role by acting as mobile base stations, particularly for demanding vehicle-to-everything (V2X) applications. 
In this scenario, one of the most challenging problems is the design of trajectories for multiple UAVs, cooperatively serving the same area. 
Such joint trajectory design can be performed using multi-agent deep reinforcement learning (MADRL) algorithms, but ensuring collision-free paths among UAVs becomes a critical challenge. 
Traditional methods involve imposing high penalties during training to discourage unsafe conditions, but these can be proven to be ineffective, whereas binary masks can be used to restrict unsafe actions, but naively applying them to all agents can lead to suboptimal solutions and inefficiencies. To address these issues, we propose a rank-based binary masking approach. Higher-ranked UAVs move optimally, while lower-ranked UAVs use this information to define improved binary masks, reducing the number of unsafe actions. This approach allows to obtain a good trade-off between exploration and exploitation, resulting in enhanced training performance, while maintaining safety constraints.
\end{abstract}

\begin{keywords}
UAVs, V2X, Reinforcement Learning, Collision-avoidance.
\end{keywords}

\section{Introduction}
\label{sec:intro}
Nowadays, the use of \acp{UAV} as flying \acp{BS}, namely \acp{UABS}, has gained considerable attention in the literature as a promising technology for future 6G networks \cite{EssayHentati,9815631}. The ability of \ac{UABS} to operate at high altitudes opens up the possibility of providing \ac{LoS} links with ground users, leading to enhanced network coverage and capacity for a wide variety of applications. An important use case is foreseen in vehicular communications~\cite{Extended_Sensing_1,white_paper}, where vehicles have to exchange large amounts of information with the network and among themselves. In such scenarios, \ac{UABS} can significantly enhance network performance.

Moreover, addressing complex urban environments poses a crucial challenge: the design of optimized trajectories for multiple \ac{UABS} to ensure a desired network performance. Additionally, introducing a coordination mechanism among \acp{UABS} becomes crucial to ensure optimal coverage of the environment while avoiding overlapping trajectories that may lead to interference or, even worse, flight collisions.


Extensive research has been conducted on collision avoidance for \ac{UAV} swarms. Conventional methods include \ac{VO}, which involves frequent changes in \acp{UAV} velocities \cite{SnaBerGuy:J11}, \ac{APF}, which does not fully leverage drone cooperation \cite{HuaLow:C18}, and \ac{PSO}, which generates energy inefficient zigzag trajectories \cite{SonWanZou:J19}. 
Addressing these challenges often requires complex mathematical models and significant computational time. In response, machine learning-based solutions \cite{10095817, 9909656, marini2022continual}, particularly \ac{MADRL}-based algorithms \cite{TesFavGio:C20, GueGuiDar:J23}, offer promising alternative approaches to mitigate such issues.
Many existing works in the literature address collision avoidance by introducing constraints on the agents' mobility models and tweaking the reward function. These constraints enforce a minimum safety distance between any two agents at any time instant \cite{WanGur:J22,XuZhaLi:J22,ThuDeg:J22,QieShiShe:J19}. 
In these works, the reward function is modified to include a simple penalty term based on mutual distances between agents.
A slightly different approach is proposed in \cite{ThuDeg:J22}, where each agent is surrounded by a ‘repulsive field’ that induces negative and positive rewards proportional to encroachment and target distances, respectively.
Nevertheless, such methods restrict the movement space of agents, that are forced to learn sub-optimal policies. 
Furthermore, solutions like the ones proposed in the aforementioned works limit the collision probability but do not entirely prevent collisions, posing challenges when applied to real-world systems.
  
This paper presents a \ac{MADRL} model that can optimize the trajectories of a swarm of \ac{UABS} deployed over an urban scenario. The model is characterized by a mask-based mechanism that allows \ac{UABS} to avoid any flight collisions, as well as possible interference with other agents that are serving vehicular users under their fields of view. 
More specifically, to improve the agents' performance, a ranking system is introduced: high-ranked agents will have the opportunity to follow the best trajectories, whereas low ones will have to respect the safety constraint using binary masking. 
Results are presented in terms of the cumulative reward collected by agents, showing the convergence property of the training algorithm designed, and in terms of the percentage of satisfied users, indicating how well the proposed system can meet the stringent requirements of \ac{V2X} applications while respecting the safety constraint introduced. 

The rest of the paper is organized as follows: Section~\ref{sec:system_model} describes the system model, Section~\ref{sec:multi_agent} offers an overview of the \ac{MADRL} model and a detailed analysis of the proposed anti-collision mechanism, Section~\ref{sec:results} shows the results obtained and Section~\ref{sec:conclusion} depicts the main conclusions.

\section{System Model} \label{sec:system_model}
\begin{figure}[t]
\centering
\includegraphics[width=\textwidth, trim={60 50 60 50}, clip]{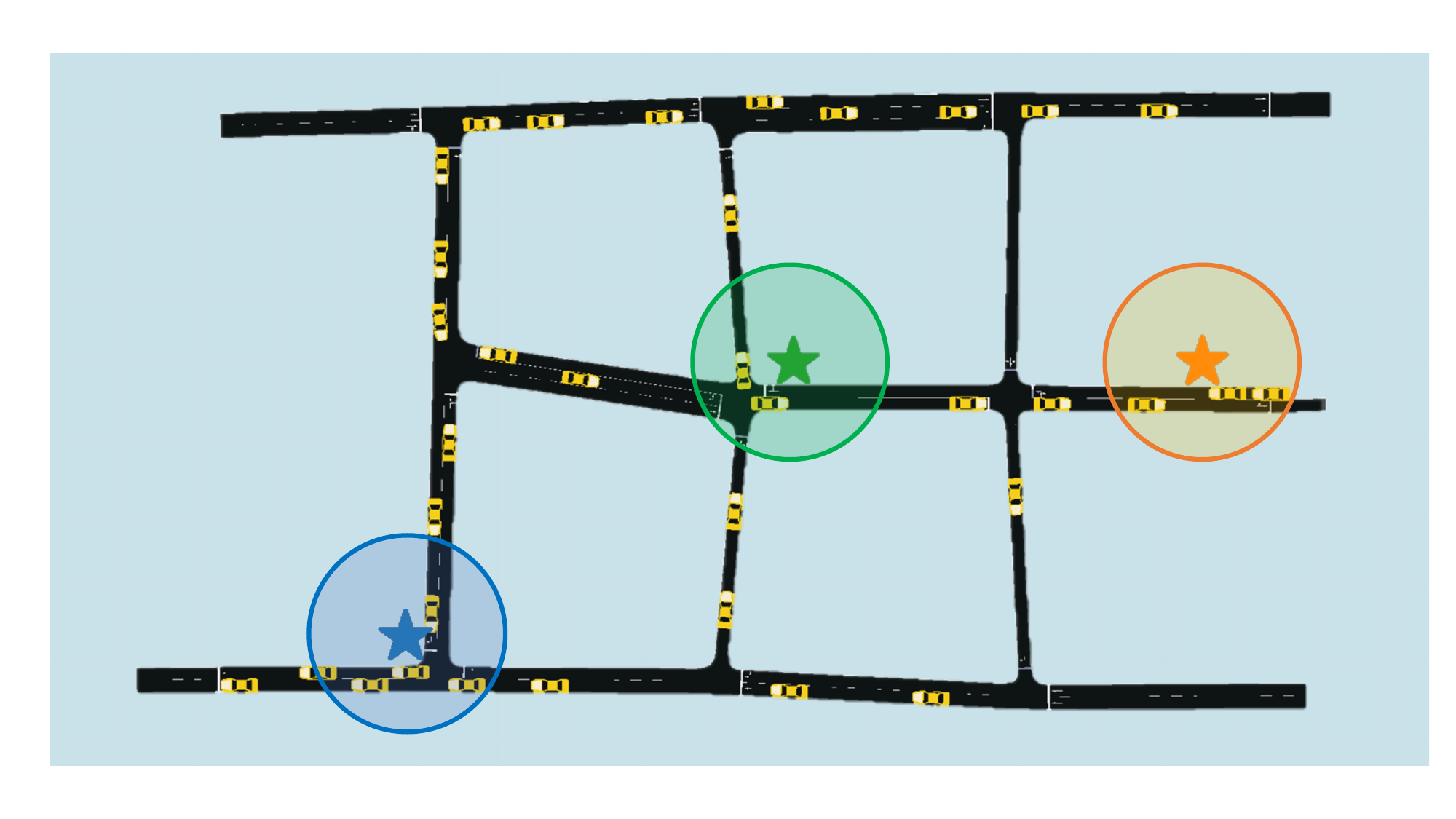}
\caption{
Manhattan-style street layout showcasing \acp{UABS} as star points, with colored circles indicating their coverage areas.
}
\label{fig:scenario}
\end{figure}
\subsection{Scenario}
Let us consider a set $\G$ of vehicles, namely \acp{GUE}, moving within an urban area while running an extended sensing application. This application necessitates each \ac{GUE} to gather data from its onboard sensors and exchange information with other \acp{GUE} and the network. 
Let us also consider a set $\U$ of $M$ \acp{UABS}, equipped with mmWave radio antenna systems working at carrier frequency $f_c$, flying in the same region and acting as relays to provide reliable and continuous service, capable of meeting stringent~requirements.

We assume that each \ac{UABS} $u\in\U$ flies at constant speed $v$, with fixed altitude $h$, with positions on xy-plane given by $(x^{(u)}_t, y^{(u)}_t)$ at each discrete time instant $t$. 
Time is discretized as $t=0, 1, \ldots, T$, where $T$ is the duration of the flight mission and each timestep has duration $\delta t$. A flight comprises a discrete sequence of positions assumed at each timestep.

It is assumed that each \ac{UABS} always has an active and reliable \ac{C2} link towards a central unit, hereafter denoted as \textit{controller}. 
We assume that the \ac{C2} links operate on distinct frequency bands w.r.t. those used for communication with \acp{GUE}, ensuring no interference between the two systems. 

The service area, depicted in Fig.~\ref{fig:scenario}, follows a Manhattan-style street layout. The routes taken by \acp{GUE} and their movement are simulated using \ac{SUMO}~\cite{sumo}. Since in this work we focus on the anti-collision mechanism, in our scenario, we simulate a small portion of a city where \acp{GUE} are concentrated.

\subsection{Application Requirements}
For this work, we focus on the collection of uplink packets, from \acp{GUE} to \acp{UABS}. 
It is assumed that \acp{GUE} transmit one packet per timestep, i.e., they generate periodic traffic with period $\delta t$.
A \ac{GUE} is \textit{served} at timestep $t$ if its packet is correctly received by at least one \ac{UABS}, i.e., the corresponding uplink \ac{SNR} $\rm SNR^{(g,u)}$ evaluated at one \ac{UABS} $u$ exceeds a threshold, $\mathrm{SNR}_{\rm th}$. 
Let us define a service window as a series of $N$ consecutive timesteps starting from a generic timestep $\bar{t}$.
We now assume that each \ac{GUE} needs to be served for at least $\hat{N_s}$ time slots inside the same service window, to provide continuous service.
Then, a \ac{GUE} is said to be \textit{satisfied} in a service window if $N_s\geq\hat{N_s}$, where $N_s$ is the number of slots for which it is served. 
With these definitions, we model the reliability and continuity of service. Furthermore, to keep track of the service history, a priority term $p_{t}^{(g)}$ is assigned to each \ac{GUE}~$g$, and for each service window varies as follows: 
\begin{numcases}{p_{t}^{(g)} =}
    1,  & if $t=\bar{t}$ \\
    p_{t-1}^{(g)} +1, & if $\exists u \in \U$ : $\rm SNR_t^{(g,u)} \geq \mathrm{SNR}_{\rm th}$ \\
    p_{t-1}^{(g)}, & otherwise.
\end{numcases}

\subsection{Channel Model} \label{sec:Channel Model}
The channel model considered is the 3GPP \ac{UMa}~\cite{TR38901}. For a generic link between two terminals, such a model defines a probability to be in \ac{LoS} condition, $\rho_L$. The \ac{SNR} in dB can be expressed as $\mathrm{SNR} = P_{\rm tx} +G_{\rm tx} + G_{\rm rx}- PL -P_{\rm n}$, where $P_{\rm tx}$ is the transmit power in~dBm, $G_{\rm tx}$ and $G_{\rm rx}$ are, respectively, the gains of the transmitter and receiver antenna in dB, $PL$ is the path loss in dB (from Table 7.4.1-1 and 7.4.2-1 in \cite{TR38901}), which depends on $\rho_L$, and $P_{\rm n}$ is the noise power at the receiver side in~dBm. 
Furthermore, we assume \acp{UABS} use beamforming generating $\B=9$ circular beam footprint on the ground in a 3x3 grid.
By defining the field of view of the \ac{UABS} on the vertical plane as $\phi$, the solid angle of a single beam can be derived as 
$\Phi_{\rm beam} \approx \frac{2\pi(1-cos(\phi/2))}{\B}$
and the maximum gain $G$ can be expressed as $G=\frac{41000}{\big(\Phi_{beam}\frac{360}{2\pi}\big)^2}$, assuming an ideal radiation pattern, with gain $G$ inside $\Phi_{\rm beam}$ and 0~dB outside \cite{beamforming_book}. 
\section{Multi Agent Reinforcement Learning Model} \label{sec:multi_agent}
\begin{figure}[t]
    \includegraphics[width=.78\columnwidth]{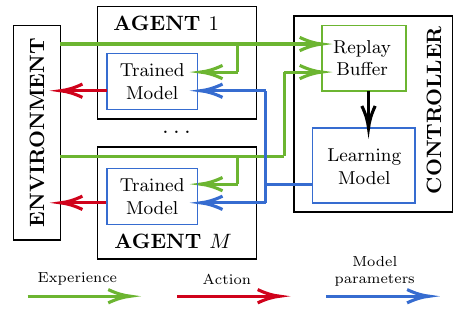}
    \caption{Multi-agent \acs{DRL} architecture. Agents continuously interact with a shared environment. Experiences are collected in a shared replay buffer at the central controller, which periodically updates the learning model and sends it back to each agent. }
    \label{fig:MADRL_architecture}
\end{figure}

The architecture of the proposed \ac{MADRL} algorithm is depicted in Fig.~\ref{fig:MADRL_architecture}. Multiple agents (i.e., the \acp{UABS}) cooperate to accomplish a common task in a shared, dynamic, and unknown environment, that is the urban area with \acp{GUE} moving along the streets. For these reasons, \acp{UABS} must interact with each other to learn how to effectively perform the task and find a behavioral policy $\pi$, represented by a \ac{NN} whose parameters must be tuned by means of a learning algorithm. The controller is a central unit that collects all the experiences of the agents in a replay buffer, trains the learning model, and periodically shares the updated policy with the agents.

At timestep $t$ each \ac{UABS} $u$ observes a limited and local representation of the environment, namely $o_t^{(u)}$.
We define a local observation as a tuple $o_t^{(u)}=(x_t^{(u)}, y_t^{(u)}, t, \mathbf{L}_t, \mathbf{b}_t^{(u)})$, where
$\mathbf{L}_t$ is a matrix containing all the current locations of agents in the fleet, 
and $\mathbf{b}_t^{(u)}$ is the \textit{per beam information}, i.e., a vector of length $\B$ whose elements $b_{i,t}^{(u)}$ corresponds to the sum of priority $p_t^{(g)}$ of \acp{GUE} under the $i$-th beam of $u$-th agent at timestep $t$. 
Based on the policy $\pi$, each \ac{UABS} selects its own action $a_t^{(u)}$ belonging to the set $\mathcal{A}$ of possible directions of movement, with $\mathcal{A} = \{ \uparrow, \rightarrow, \downarrow, \leftarrow \}$.
As a consequence of the collective actions taken by all agents, the environment will change 
leading to new local observations $o_{t+1}^{(u)}$ for each agent.
Local observations and chosen actions are sent back to the controller so that a per agent reward $r_{t}^{(u)}$ can be computed.
The reward indicates how favorable or unfavorable has been choosing action $a_t^{(u)}$ within the context of local observations $o_t^{(u)}$ and $o_{t+1}^{(u)}$. 
At this point, the experience tuple $(o_t^{(u)}, a_t^{(u)}, r_{t}^{(u)}, o_{t+1}^{(u)})$ is stored in the shared replay buffer. 
By defining $d_{t}^{(u,w)}$ as the euclidean distance between agents $u$ and $w$ at a timestep $t$ and $\lVert \mathbf{x} \rVert_{1}=\sum_{x \in \mathbf{x}} |x|$ the 1-norm of a vector $\mathbf{x}$, the reward function can be written as:
\begin{subnumcases}
{r_{t}^{(u)} =}\label{eq:reward}
    \lVert \mathbf{b}_{t+1}^{(u)} \rVert_{1}, & if $\forall w \in \U\setminus \{u \}$ : $d_{t+1}^{(u,w)}\ge d_\text{th}$ \label{eq:reward_priority} \\
    -\lambda_{s},& if $\exists w \in \U\setminus \{u \}$ : $d_{t+1}^{(u,w)} < d_\text{th}$ \label{eq:distance_penalty} \\
    -\lambda_{c},& if $\exists w \in \U\setminus \{u \}$ : $d_{t+1}^{(u,w)} < 1$. \label{eq:collision_penalty}
\end{subnumcases}
\eqref{eq:reward_priority} corresponds to the sum of the priority of \acp{GUE} covered by the beams of \ac{UABS} $u$, after choosing action $a_t^{(u)}$ at time~$t$, reaching a new position and getting the per beam information at timestep~$t+1$. 
On the other hand, when distance requirements are not met
in \eqref{eq:distance_penalty} and \eqref{eq:collision_penalty}, agents get penalties equal to $\lambda_s$ and $\lambda_c$, respectively.
The former condition happens when two agents start to interfere with each other, with $d_{\rm th}=2h\arctan{\phi}$ being twice their coverage radius, whereas the latter corresponds to a collision event between them. 

After $T_u$ timesteps, the controller will update the \ac{NN} parameters following a \ac{3DQN} learning algorithm and share the new model with each agent. 
To reduce complexity, improve performance, and leverage a \textit{centralized learning distributed execution} paradigm, the trained policy $\pi$ is shared among all agents so that each one exploits the same one. Additionally, centralized training avoids high consumption of energy at the \ac{UABS}, whereas the distributed execution, guaranteed by the local model replica, allows each \ac{UABS} to act quickly once new observations arrive, reducing possible delays.

Fundamental concepts of the \ac{3DQN} implementation are reported hereafter. More details can be found in~\cite{10095817}.
In short, the aim of \ac{3DQN} algorithm is to estimate the Q-values, representing the expected future rewards when choosing an action $a$ in a given observation $o$, for all possible actions and observations. 
Periodical updates of the \ac{NN} parameters are based on the stochastic gradient descent on mini-batches of local experiences $(o_t, a_t, r_t, o_{t+1})$ randomly sampled from the shared replay buffer. 
These experiences can be collected either by \textit{exploiting} the current policy $\pi$ and choosing the action with the highest Q-value when observation $o_t$ is perceived, or by \textit{exploring} the environment, thus choosing random actions, possibly transitioning to new positions obtaining higher rewards, and learn a better policy. Agents must strike a good balance between exploration and exploitation by following an $\epsilon$-greedy policy.
Additionally, a \textit{double} loss calculation, to avoid overestimation, and a \textit{dueling} architecture, useful to increase the sample-efficiency of the algorithm, are exploited to provide better learning outcomes. Indeed, \ac{3DQN} provides good performance in this type of scenario, balancing training convergence and real-time execution of new trajectories. 

\subsection{Anti-Collision Mechanism}

In real-world scenarios, the likelihood of \acp{UABS} flying in dangerously close proximity is not negligible, making it imperative to effectively diminish the collision risk during their missions.  
The widely-adopted penalty-based strategy \cite{WanGur:J22,XuZhaLi:J22,ThuDeg:J22,QieShiShe:J19} requires
the definition of a reward function that imposes penalties on agents when they get too close, as outlined in \eqref{eq:distance_penalty} and \eqref{eq:collision_penalty}. As simulations will show, this approach often results in agents colliding with each other, failing to provide a definitive solution to the problem.
Hence, we introduce two novel collision avoidance mechanisms, namely \textit{flat} and \textit{rank masking}, which rely on constraining the agents' action space to entirely avert collisions.

The flat masking algorithm works as follows:
i) at timestep~$t$, the controller examines each agent and assesses all their potential future positions;
ii) for each couple of agents $u,w \in \U$, actions for which $d_{t+1}^{(u,w)} < d_{\rm th}+v \, \delta t$ are classified as illegal and removed from the action space, resulting in temporarily restricted sets of safe actions $\Tilde{\mathcal{A}}_t^{(u)}$ and $\Tilde{\mathcal{A}}_t^{(w)}$. 
The term $v \, \delta t$ is introduced as an additional safety factor accommodating for the uncertainty caused by the selection of random actions that are not known in advance by the agents. 
The restricted binary masks are sent to each agent via the~\ac{C2}~link.

It becomes evident that the utilization of flat masking significantly dampens the agents' capacities for both exploration and exploitation. For instance, in a scenario where many agents are directed toward a specific location, the masking effect induces a sense of mutual repulsion, preventing any of them from reaching the destination. This directly affects the quality of service provided to \acp{GUE}. 

To address this limitation and further improve the system's performance, we introduce the novel concept of rank masking:
i) a rank is assigned to each agent by means of a score function $F(u)=\lVert \mathbf{b}_{t}^{(u)} \rVert_{1}$, that is the sum of the priorities of \acp{GUE} currently covered by the $u$-th agent; 
ii) the couples $u,w \in \U$ for which $d_{t}^{(u,w)} < d_{\rm th} + 2 \, v \, \delta t$ are identified as potential colliders;
iii) if two agents can potentially collide, the one with the highest rank is free to take the next action without any limitation, according to its policy;
iv) the agent $u$ with the lowest rank calculates its constrained set of actions, $\Tilde{\mathcal{A}}_t^{(u)}$, for which $d_{t+1}^{(u,w)} < d_{\rm th}$, based on the action chosen by agent $w$. 

\section{Results} \label{sec:results}
In this section, we provide simulation results to compare penalty and masking-based collision avoidance mechanisms, proving that the former is less efficient than the latter in terms of performance. It is also shown that ranked-based masking performs better than flat masking. As performance metrics, we used the cumulative reward, i.e., the sum of agents' reward, $R = \sum_{u \in \U}\sum_{t=0}^{T} r_t^{(u)}$, which provides feedback on the training performance of the designed \ac{MADRL} algorithm, and the percentage of satisfied users $P_g$, calculated as 
\begin{equation}
    P_g = \frac{1}{\vert \G \vert} \sum_{g \in \G} \frac{N_{g}^{(sat)}}{N_g}
\end{equation}
where $N_g$ is the total number of service windows of the \mbox{$g$-th}~\ac{GUE}, while 
$N_{g}^{(\rm sat)}$ are those for which $N_s \ge \hat{N_s}$.
The simulation parameters used are reported in Table \ref{tab:sim_param}.

$M$ agents are trained for $N_t$ episodes of duration $T$. One \ac{UABS} is randomly placed inside the area, while the others are uniformly distributed on its perimeter. The scenario considered has dimensions $350\times170 \hspace{0.2em} \rm m^2$.
Every $N_e$ training episodes we perform an evaluation of the current policy $\pi$ on vehicle traces that have not been used for the training. 
During evaluation agents are not allowed to perform random actions, so we can evaluate and compare their behavior in a fixed and fair setting.


\begin{table}[t]
\centering
\caption{Simulation parameters}
\label{tab:sim_param}
\vspace{-.1cm}
\small
\begin{tabularx}{\linewidth}{|Y|c|Y|c|Y|c|}
\hline
$ M $ & $3$ & $h$ & $100$ m & $v$  & 20~m/s \\
\hline
$\phi$ & 40°  & $|\mathcal{G}|$ & $80$ & $\delta t$ & 1~s \\
\hline
$f_c$ & $30$~GHz & $P_{\rm tx}$ & 14 dBm & $P_{\rm n}$ & -106.4 dBm \\
\hline
$G_{\rm tx}$ & 0 dB &  $G_{\rm rx}$, & 38 dB & $\mathrm{SNR}_{\rm th}$ & -13.7 dB\\
\hline
$T$ & $80$ s & $\lambda_{s}$ & 10 & $\lambda_{c}$ & 1000  \\
\hline
$N_{t}$ & 1000 & $N_{e}$ & 20 & $T_u$ & 1\\
\hline
\end{tabularx}
\end{table}  
\begin{table}[t]
\centering
\caption{Average number of collisions occurrences}
\label{tab:collision_count}
\vspace{-.1cm}
\small
\begin{tabularx}{\linewidth}{c|c|Y|Y|}
\cline{2-4}
 & \multirow{2}{0.9cm}{Penalty} & Flat Masking & Rank Masking \\ \hline
\multicolumn{1}{|l|}{Collisions in Training} & $170$ $(17$\%$)$ & $0$ $(0$\%$)$ & $0$ $(0$\%$)$ \\ \hline
\multicolumn{1}{|l|}{Collisions in Evaluation} & $14$ $(28$\%$)$ & $0$ $(0$\%$)$ & $0$ $(0$\%$)$ \\ \hline
\end{tabularx}%
\end{table}

\begin{figure}[t]
\centering
\includegraphics[width=\textwidth, trim={10 0 0 0}, clip]{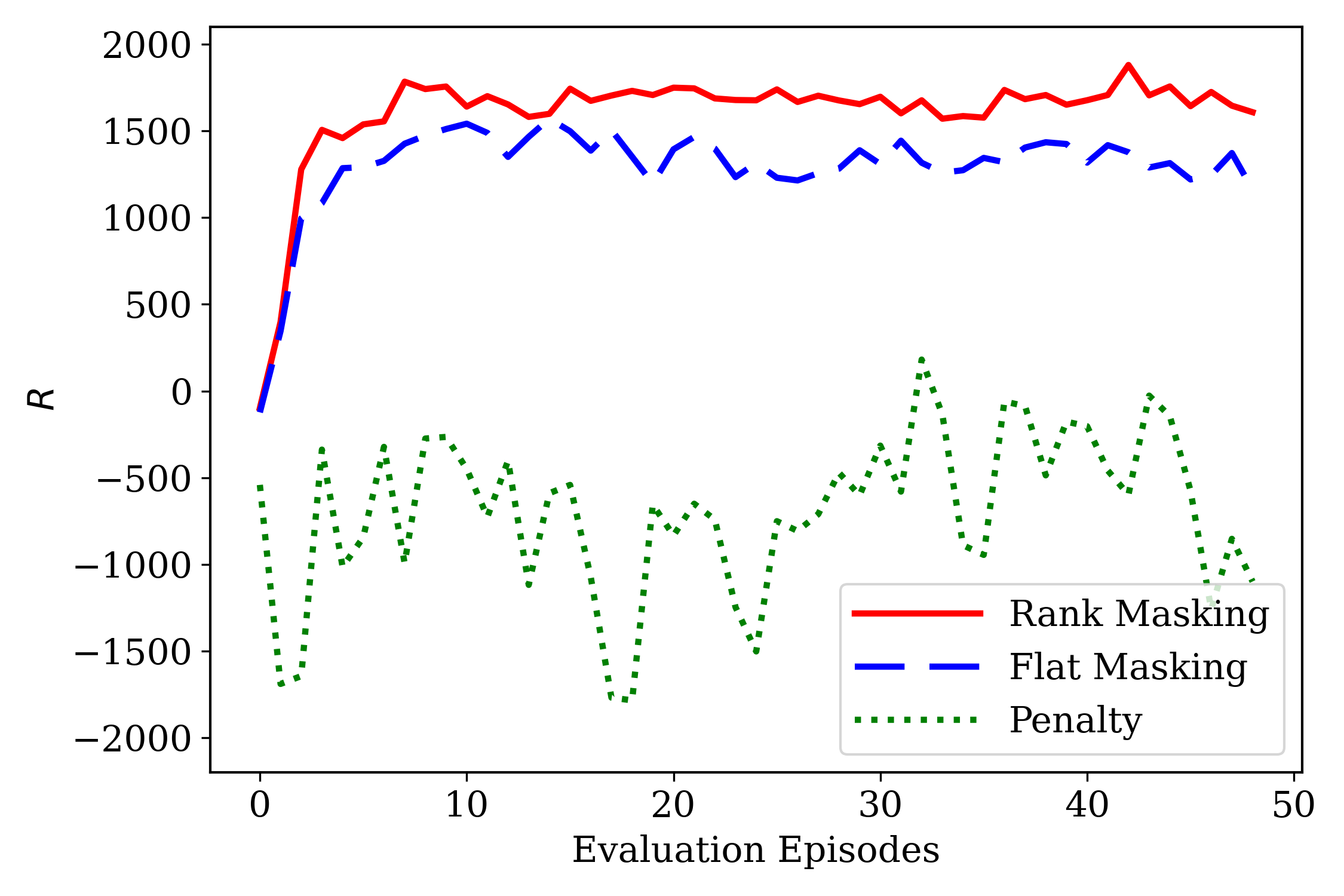}
\caption{Cumulative reward in evaluation episodes for different anti-collision mechanisms.}
\label{fig:evaluation}
\end{figure}

Table~\ref{tab:collision_count} lists the average number of collisions in training and evaluation episodes, showing how masking can be used to completely avoid collisions in contrast to penalty-based algorithms.

In Fig.~\ref{fig:evaluation} the cumulative reward, $R$, obtained during evaluation episodes is shown.
It is possible to notice that the use of the penalty-based approach turns out to be ineffective in this highly dynamic scenario due to the mix of penalties received by the agents.
Additionally, thanks to the improved exploration and exploitation capability, rank-based masking allows agents to collect higher rewards than flat masking.
Let us remark that when anti-collision mechanism is active the agents only collect the positive rewards outlined in~\eqref{eq:reward}. 


\begin{figure}[ht]
\centering
\includegraphics[width=0.95\textwidth, trim={5 10 9 10}, clip]{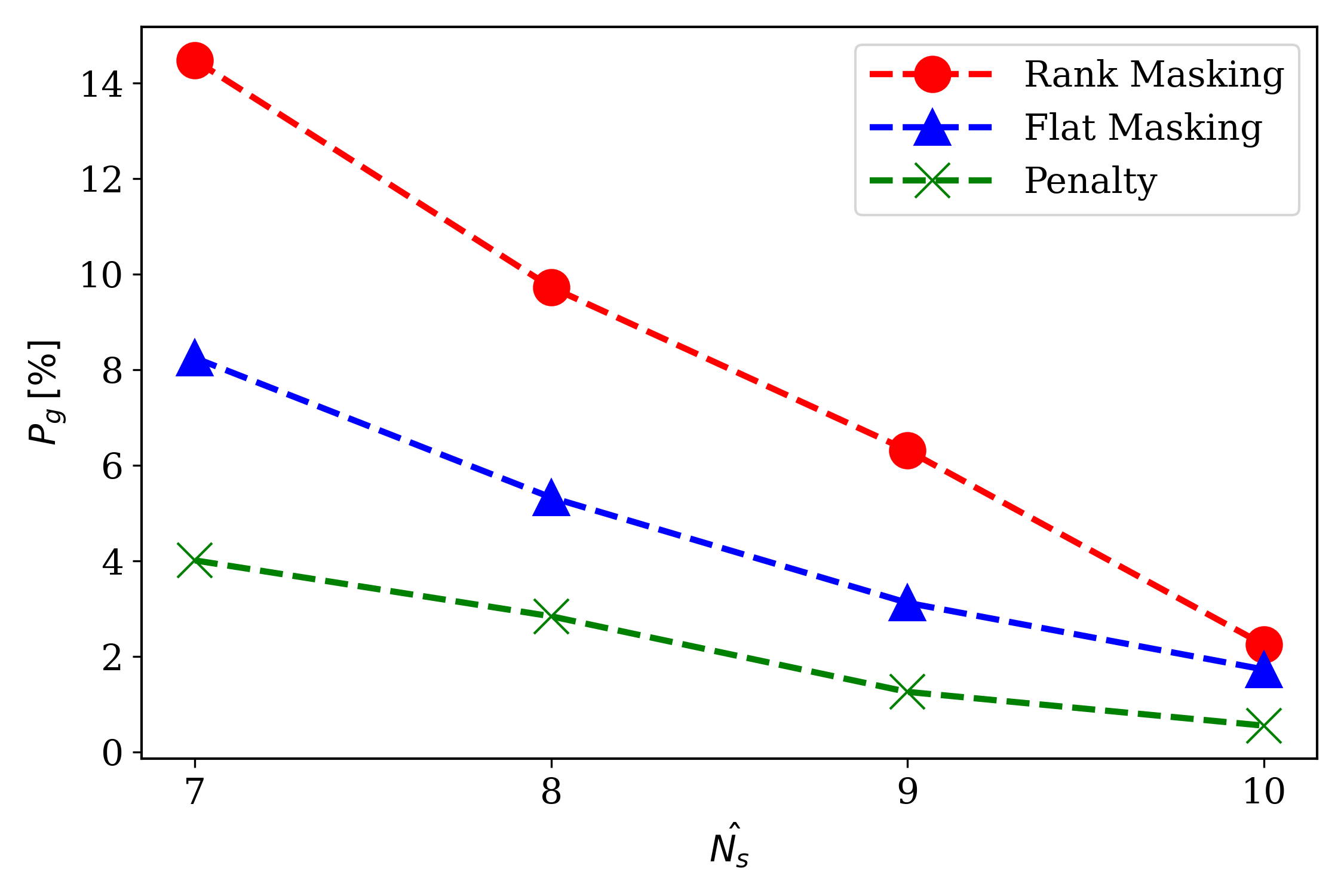}
\caption{Percentage of satisfied users for different anti-collision mechanisms.}
\label{fig:pg}
\end{figure}
Fig.~\ref{fig:pg} shows the $P_g$ by varying the service threshold, $\hat{N_s}$. Values are obtained from the best evaluation episode for each setting. 
As expected, rank-based masking is again proven to be better than flat masking. However, both masking strategies easily outperform the trajectories obtained with penalty-based training. 
It is worth mentioning that the low values of user satisfaction are mainly attributed to the \acp{UABS}' limited coverage footprint and the beam ideality hypothesis, causing a maximum instantaneous coverage area that is only $12\%$ of the overall service area.
Indeed, the system performance can be improved by introducing a fixed ground network capable of serving \acp{GUE} cooperating with \acp{UABS}. 
\section{Conclusions} \label{sec:conclusion}
This paper addresses anti-collision methods for a group of autonomous \ac{UABS} whose aim is to continuously serve groups of \acp{GUE}. 
Indeed, when trajectories are learned using \ac{MADRL} algorithms, it becomes difficult to respect basic safety constrain on the \acp{UABS}' mutual distance without a proper design.
To this end, three strategies for collision avoidance are presented and compared. The study finds that incorporating penalties into the reward function is not enough to avoid collisions, necessitating a binary mask to prevent unsafe actions.
Moreover, a ranking system can be adopted to ensure service continuity and improve the exploration and exploitation capabilities of the fleet during the training phase.

\section {Acknowledgment}
This work has been carried out in the framework of the CNIT WiLab-Huawei Joint Innovation Center and also supported by the European Union under the Italian National Recovery and Resilience Plan (NRRP) of NextGenerationEU, partnership on “Telecommunications of the Future” (PE00000001 - program “RESTART”, Structural Project 6GWINET).
We thank Aman Jassal for the very fruitful discussion on this paper.
\acrodef{3GPP}{3rd Generation Partnership Project}
\acrodef{5G}{5-th Generation}
\acrodef{APF}{artificial potential field}
\acrodef{BS}{Base Station}
\acrodef{CAM}{Cooperative Awareness Message}
\acrodef{CPM}{Collective Perception Message}
\acrodef{DDPG}{Deep Deterministic Policy Gradient}
\acrodef{DQN}{Deep Q-Learning}
\acrodef{DDQN}{Double Deep Q-Learning}
\acrodef{ESN}{Echo State Network}
\acrodef{IoT}{Internet of Things}
\acrodef{GPS}{global positioning system}
\acrodef{GUE}{Ground User Equipment}
\acrodef{LoS}{Line-of-Sight}
\acrodef{MADRL}{Multi-Agent Deep Reinforcement Learning}
\acrodef{ML}{Machine Learning}
\acrodef{NFZ}{No Fly Zone}
\acrodef{NLoS}{Non Line-of-Sight}
\acrodef{NN}{Neural Network}
\acrodef{DNN}{Deep Neural Network}
\acrodef{PSO}{particle swarm optimization}
\acrodef{RL}{Reinforcement Learning}
\acrodef{RSSI}{Received Signal Strength Indicator}
\acrodef{SAR}{Search and Rescue}
\acrodef{SNR}{Signal-to-Noise Ratio}
\acrodef{SUMO}{Simulation of Urban MObility}
\acrodef{UABS}{Unmanned Aerial Base Station}
\acrodef{UAV}{Unmanned Aerial Vehicle}
\acrodef{UE}{User Equipment}
\acrodef{UMa}{Urban Macro}
\acrodef{V2X}{Vehicle-To-Everything}
\acrodef{VO}{velocity obstacles}
\acrodef{QoE}{Quality of Experience}
\acrodef{FFS}{Finite Fourier Series}
\acrodef{DRL}{Deep Reinforcement Learning}
\acrodef{O2O}{Outdoor-to-Outdoor}
\acrodef{O2I}{Outdoor-to-Indoor}
\acrodef{MDP}{Markov Decision Process}
\acrodef{AoI}{Age of Information}
\acrodef{3DQN}{Double Dueling Deep Q Network}
\acrodef{OHE}{One Hot Encoding}
\acrodef{C2}{Command and Control}
\bibliographystyle{IEEEtran}
\bibliography{IEEEabrv, main}

\end{document}